\documentclass[letterpaper, 10 pt, conference]{ieeeconf} 
\IEEEoverridecommandlockouts 
\usepackage{graphicx}
\usepackage{balance} 
\usepackage{comment}
\usepackage{colortbl}
\usepackage{subfig}
\usepackage{amssymb}
\usepackage{amsmath}

\usepackage{fancyhdr}
\usepackage{ragged2e}
\usepackage{algorithm}
\usepackage{algpseudocode}

\def\BibTeX{{\rm B\kern-.05em{\sc i\kern-.025em b}\kern-.08em
    T\kern-.1667em\lower.7ex\hbox{E}\kern-.125emX}}
\begin{document}

\fancypagestyle{firstpage}
{
    \fancyhf{}

    \fancyhead[L]{\small \centering \noindent Extended abstract accepted at the 2nd RL-CONFORM Workshop at IEEE/RSJ IROS'22 Conference, Kyoto, Japan, 2022. 
    }

}

\title{\LARGE \bf
Broad-persistent Advice for Interactive\\Reinforcement Learning Scenarios
}

\author{
Francisco Cruz$^{1}$,
Adam Bignold$^{2}$,
Hung Son Nguyen$^{3}$,
Richard Dazeley$^{3}$, and
Peter Vamplew$^{2}$
\thanks{$^{1}$Francisco Cruz is with the School of Computer Science and Engineering, University of New South Wales, Sydney, Australia.
        {\tt\small f.cruz@unsw.edu.au}}%
\thanks{$^{2}$Adam Bignold and Peter Vamplew are with the School of Engineering, IT and Physical Sciences, Federation University, Ballarat, Australia.
        {\tt\small \{a.bignold, p.vamplew\}@federation.edu.au}}%
\thanks{$^{3}$Hung Son Nguyen and Richard Dazeley are with the School of Information Technology, Deakin University, Geelong, Australia.
        {\tt\small \{hsngu, richard.dazeley\}@deakin.edu.au}}%
}

\maketitle

\begin{abstract}
The use of interactive advice in reinforcement learning scenarios allows for speeding up the learning process for autonomous agents. 
Current interactive reinforcement learning research has been limited to real-time interactions that offer relevant user advice to the current state only. 
Moreover, the information provided by each interaction is not retained and instead discarded by the agent after a single use. 
In this paper, we present a method for retaining and reusing provided knowledge, allowing trainers to give general advice relevant to more than just the current state. 
Results obtained show that the use of broad-persistent advice substantially improves the performance of the agent while reducing the number of interactions required for the trainer.

\end{abstract}


\thispagestyle{firstpage}

\section{Introduction}

Reinforcement learning (RL) is a method used for robot control in order to learn optimal policy through interaction with the environment, through trial and error~\cite{sutton2018reinforcement}. 
The use of RL in previous research shows that there is great potential for using RL in robotic scenarios~\cite{cruz2017agent, cruz2018action}. 
Especially, deep RL (DRL) has also achieved promising results in manipulation skills~\cite{Wang2020, Nguyen2019}, and on how to grasp as well as legged locomotion~\cite{Ibarz2021}. 
However, there is an open issue relating to the performance in both RL and DRL algorithms, which is the excessive time and resources required by the agent to achieve acceptable outcomes~\cite{millan2021robust, ayala2019reinforcement}. 
The larger and more complex the state space is, the more computational costs will be spent to find the optimal policy. 

\begin{figure}
\centering
\includegraphics[width=1\linewidth]{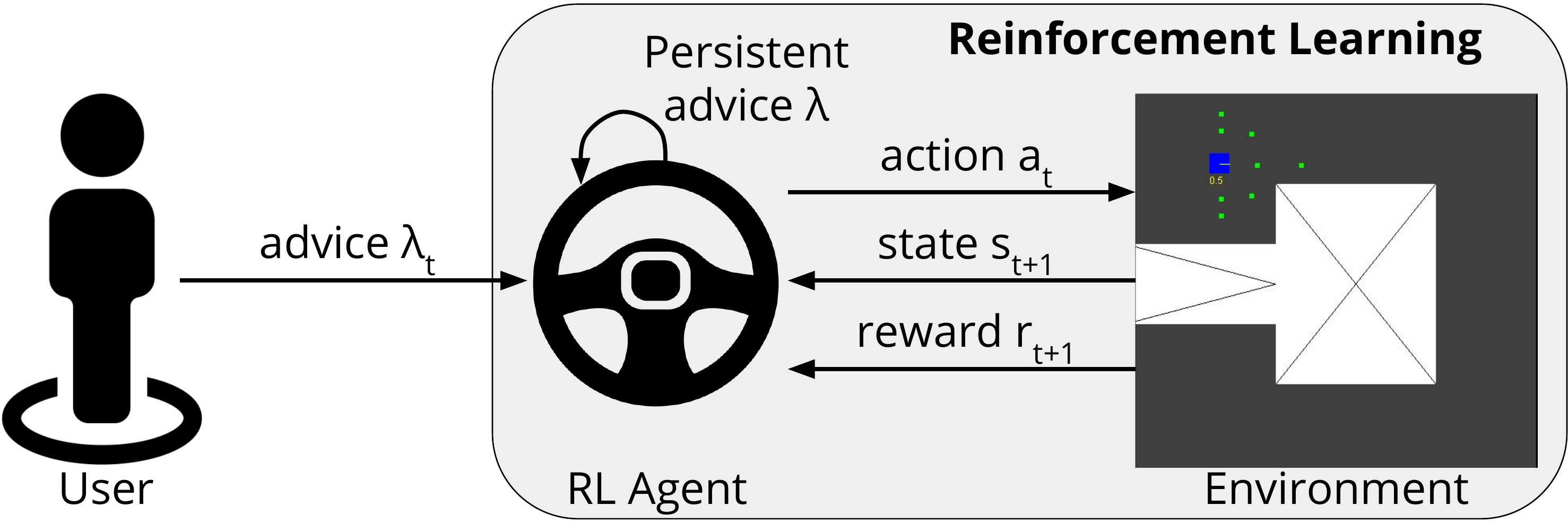}
\caption{Interactive reinforcement learning framework.
In traditional RL an agent performs an action and observes a new state and reward.
In the figure, the environment is represented by the simulated self-driving car scenario and the RL agent may control the direction and speed of the car.
IntRL adds advice from a user acting as an external expert in certain situations.
Our proposal includes the use of broad-persistent advice in order to minimize the interaction with the trainer.
}
\label{fig:irl}
\end{figure}

In this regard, interactive RL (IntRL) allows for speeding up the learning process by including a trainer to guide or evaluate a learning agent's behavior~\cite{cruz2014improving, arzate2020survey}. 
The assistance provided by the trainer reinforces the behavior the agent is learning and shapes the exploration policy, resulting in a reduced search space~\cite{bignold2022human}. 
Figure~\ref{fig:irl} depicts the IntRL approach. 
Current IntRL techniques discard the advice sourced from the human shortly after it has been used~\cite{knox2009interactively, bignold2021conceptual}, increasing the dependency on the advisor to repeatedly provide the same advice to maximize the agent’s use of it.

Moreover, current IntRL approaches allow trainers to evaluate or recommend actions based only on the current state of the environment~\cite{policyshaping, tamer}. 
This constraint restricts the trainer to providing advice relevant to the current state and no other, even when such advice may be applicable to multiple states~\cite{taylor2014reinforcement}. 
Restricting the time and utility of advice affect negatively the interactive approach in terms of creating an increasing demand on the user's time, instead of withholding potentially useful information for the agent~\cite{lin2020review}.

This work presents a broad-persistent advising (BPA) approach for IntRL to provide the agent with a method for information retention and reuse of previous advice from a trainer. 
This approach includes two components: generalization and persistence.  
Agents using the BPA approach exhibit better results than their non-using counterparts and with a substantially reduced interaction count.

\section{Broad-persistent advice}

Recent studies~\cite{bignold2021persistent, nguyen2021broad} suggest permanent agents that record each interaction and the circumstances around particular states. 
The actions are taken again when the same conditions are met in the future. 
As a consequence, the recommendations from the advisor are used more effectively, and the agent's performance improves. 
Furthermore, as there is no need to provide advice for each repeated state, less interaction with the advisor is required. 

\begin{figure}
\centering
\includegraphics[width=0.9\linewidth]{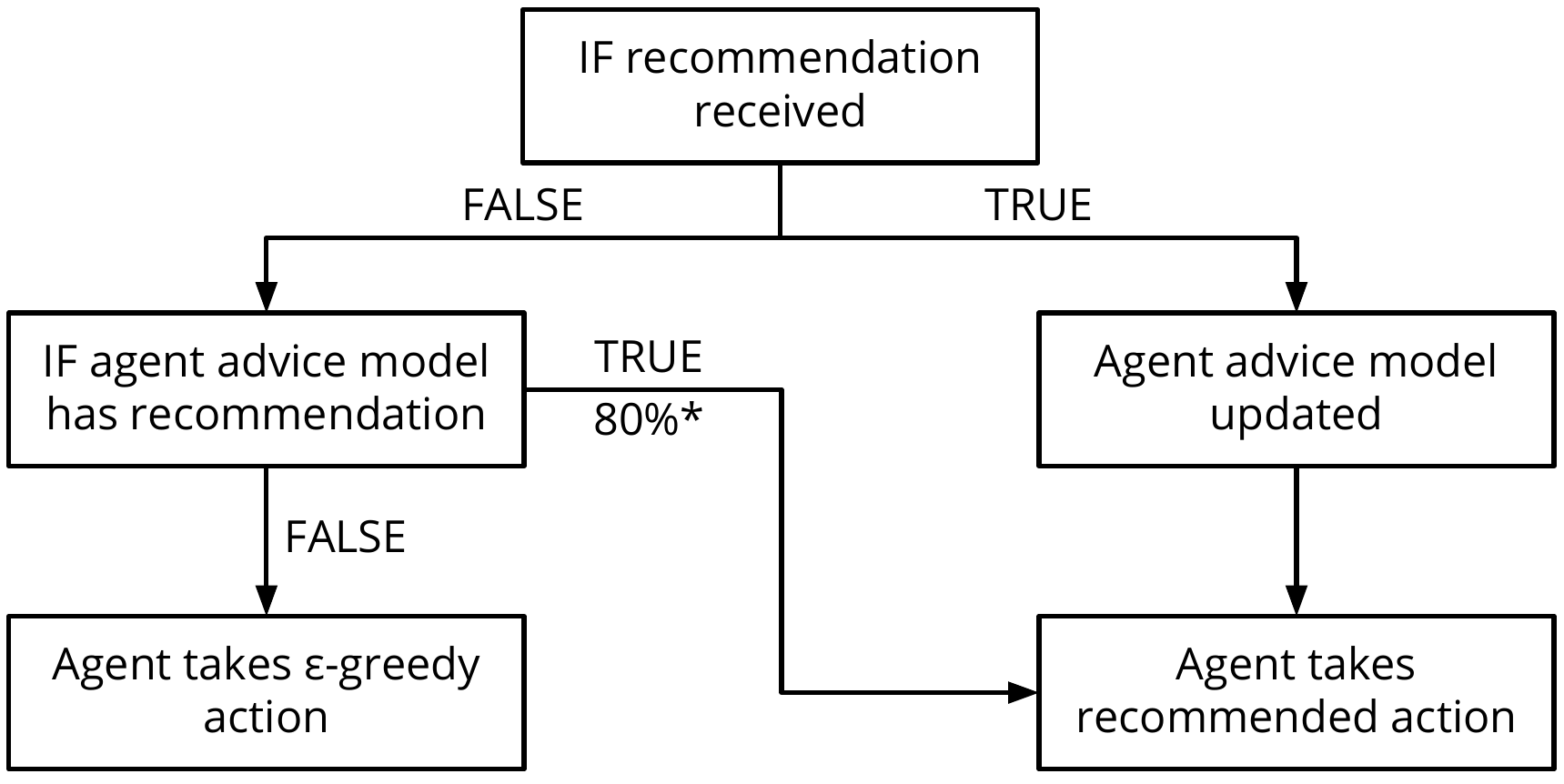}
\caption{Probabilistic policy reuse (PPR) for an IntRL agent using informative advice. If the user recommends an action on the current time step then the agent's advice model updates and the action is performed. If the user does not provide advice on the current time step, then the agent will follow previously obtained advice 80\% of the time (*decays over time) and its default exploration policy the remaining time.}
\label{fig:c7_ppr}
\end{figure}

However, as inaccurate advice is also possible, after a certain amount of time, a mechanism for discarding or ignoring advice is needed. 
Probabilistic policy reuse (PPR) is a strategy for improving RL agents that use advice~\cite{fernandez2006probabilistic}. 
Where various exploration policies are available, PPR uses probabilistic bias to decide which one to choose, with the intention of balancing between random exploration, the use of a guideline policy, and the use of the existing policy.
The PPR approach is shown in Figure~\ref{fig:c7_ppr}.

To use PPR, we need a system to store the used pairs of state-action. 
When the agent arrives at a certain state at a time step, agents using PPR need to check with the system if this state has been suggested by the trainer in the past. 
If there is advice in the memory of the model, the agent can use the option to reuse the action. 
However, there is a problem when using PPR in large and continuous domains. 
It is not efficient to build a system that stores unlimited state-action pairs. 
In addition, when the amount of state becomes too large in space, the possibility that agents revisit exactly the same state will be very small. 
Therefore, building this model will become cumbersome and inefficient in large spaces. 

BPA includes a model for clustering states and then building a system for cluster-action pairs instead of traditional state-action pairs. 
When the agent receives current state information from the environment and it does not receive any advice from the trainer, the agent will use PPR by injecting the state into the generalization model and defining its cluster. 
Then proceed to consider whether any advice pertains to the current cluster. 
If there is an action recommended in the past, the agent can reuse it with PPR selection probability, or use the default action selection policy.

\section{Experimental scenarios}

\subsection{Mountain car}
The mountain car is a control problem in which a car is located on a unidimensional track between two steep hills.
This environment is a well-known benchmark in the RL community, therefore, it is a good candidate to initially test our proposed approach.

The car starts at a random position at the bottom of the valley ($-0.6<x<0.4$) with no velocity ($v=0$).
The aim is to reach the top of the right hill.
However, the car engine does not have enough power to claim to the top directly and, therefore, needs to build momentum moving toward the left hill first.

An RL agent controlling the car movements observes two state variables, namely, the position $x$ and the velocity $v$.
The position $x$ varies between -1.2 and 0.6 in the x-axis and the velocity $v$  between -0.07 and 0.07.
The agent can take three different actions: accelerate the car to the left, accelerate the car to the right, and do nothing.

The agent receives a negative reward of $r=-1$ each time step, while no reward is given if a hill is reached ($r=0$).
The learning episode finishes in case the top of the right hill is climbed ($x=0.6$) or after 1,000 iterations in which case the episode is forcibly terminated.

\subsection{Self-driving car}
\label{c4:self_driving_car}
\begin{figure}
\centering
\includegraphics[width=0.6\linewidth]{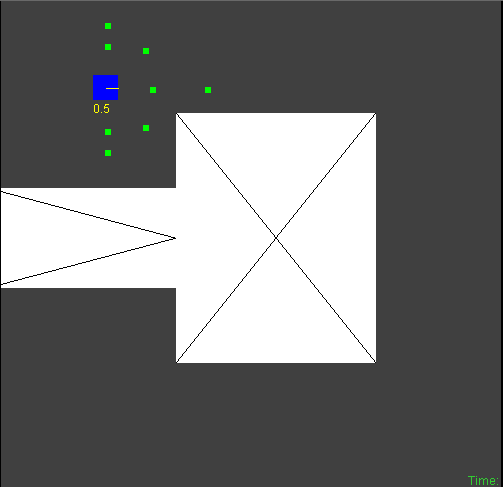} 
\caption{A graphical representation of the simulated self-driving car. 
The blue square at the top left is the car. 
A yellow line within the car indicates the current direction and the number below (in yellow) is the current velocity. 
The small green squares surrounding the car are collision sensors and will always align with the car's current direction. 
The large white rectangles are obstacles.}
\label{fig:sim_car_environment}
\end{figure}

The simulated self-driving car environment is a control problem in which a simulated car, controlled by the agent, must navigate an environment while avoiding collisions and maximizing speed. 
The car has collision sensors positioned around it which can detect if an obstacle is in that position, but not the distance to that position. 
Additionally, the car can observe its current velocity. 
All observations made by the agent come from its reference point, this includes the obstacles (e.g., there is an obstacle on my left) and the car's current speed. 
The agent cannot observe its position in the environment. 
Figure~\ref{fig:sim_car_environment} shows a representation of the environment.

In each step, the environment provides the agent reward equal to its current velocity. 
A penalty of -100 is awarded each time that the agent collides with an obstacle. 
Along with the penalty reward, the agent's position resets to a safe position within the map, velocity resets to the lower limit, and the direction of travel is set to face the direction with the longest distance to an obstacle. 

There are five possible actions for the agent to take within the self-driving car environment. 
These actions are accelerate, decelerate, turn left, turn right, and do nothing

The self-driving car environment has nine state features, one for each of the collision sensors on the car, and the current velocity of the car. 
The collision sensor state features are Boolean, representing whether they detect an obstacle at their position. 
The velocity of the agent has nine possible values, the upper and lower limits, plus every increment of 0.5 value in between. 
With the inclusion of the five possible actions, this environment has $11520$ state-action pairs.

\subsection{Simulated robot environment}
\begin{figure}
    \centering
    \includegraphics[width=0.6\linewidth]{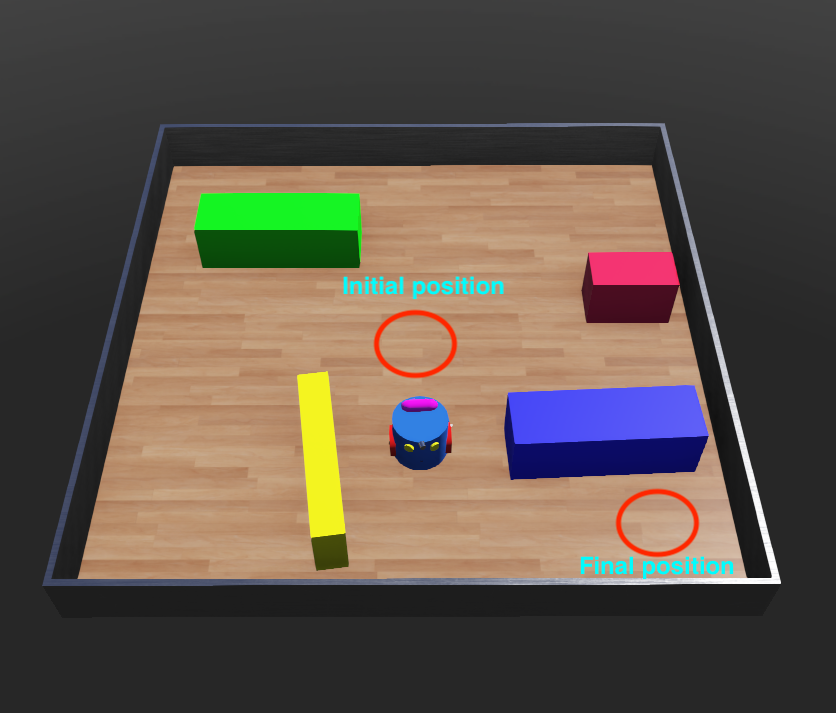}
    \caption{An example of Webots environment with initial position and final position. The robot has the goal to go from the initial position to the final position while avoiding obstacles. The robot will be returned to its initial position after any collision. }
    \label{fig:42}
\end{figure}

Additionally, we also build an environment for domestic robots using Webots, given the overall good performance shown previously~\cite{ayala2020comparison}. 
In this environment, the goal is to train the robot to go from the initial position to the target position. 
Figure~\ref{fig:42} denotes a graphic of our experimental environment in Webots.

The robot is equipped with distance sensors on its left and right eyes. 
The robot is completely unaware of its current position in the environment. 
The robot can only choose one of three actions: go straight at 3m/s, turn left, or turn right. At each step, the robot will be deducted 0.1 points if it uses the action of turning left or right, no points will be deducted if it chooses to go straight. 
This is to optimize the robot's straight movement and avoid the robot running in circles by turning left or right continuously. 
The robot is equipped with a few touch sensors next to it, to detect the collision with the environment. 
The robot will be returned to its initial position and receive 100 penalty points every time it collides on the way. 
The robot does not know where the touch sensor is located relative to itself, the only information it receives is whether it is a collision with obstacles or not. 
When the robot goes to the finish position located in the lower right corner of the environment, the robot is considered to complete the task and be rewarded with 1000 points.

The state is represented as a $64 \times 64$ RGB image taken from the top of the environment.
Three actions are allowed: go straight at 3m/s, turn left, or turn right, and the reward function is defined as turn left, right: -0.1; go straight: 0; collision: -100; reach to final position: 1000.

\section{Experimental results}

\subsection{Simulated users}

To compare agent performance, information about interactions, agent steps, rewards, and interactions are recorded. 
To identify the efficiency of BPA, we need to test the experiment with three cases: No interactive action, interactive actions without BPA, and interactive actions with BPA.

To provide feedback, we employ simulated users~\cite{bignold2021evaluation}.
Each use case of the simulated user will have different advice's accuracy and frequency. 
Frequency is the availability of the interaction of the advisor at the given time step. 
The higher frequency, the advisor has more rate for giving advice to the agent. 
Accuracy is a measure of the precision of advice provided by an advisor. 
When the advisor's accuracy is high, the action would be proposed precisely as per the advisor's knowledge. 
On the contrary, the action proposed is different from the advisor's knowledge. 
The frequency and accuracy were simulated using data from a previous user study~\cite{bignold2022human}.
These values are described in Table~\ref{c7:tab:sim_user_settings}. 

\begin{table}
\caption{Simulated users modeled for the experimental setup. 
Accuracy and availability are set using previous results obtained in a human trial as reference~\cite{bignold2022human}.
}

\begin{center}
\begin{tabular}{|l|l|l|}
 \hline
 \textbf{Name}& \textbf{Accuracy} & \textbf{Availability} \\
 \hline
 Informative Optimistic   		& 100\%    & 100\%       \\
 Informative Realistic			& 94.870\% & 47.316\%    \\
 Informative Pessimistic		& 47.435\% & 23.658\%    \\
 \hline
\end{tabular}
\end{center}
\label{c7:tab:sim_user_settings}
\end{table}

\subsection{Results}

\begin{figure}
\centering
\subfloat[Non-persistent informative]{\includegraphics[width=0.5\linewidth]{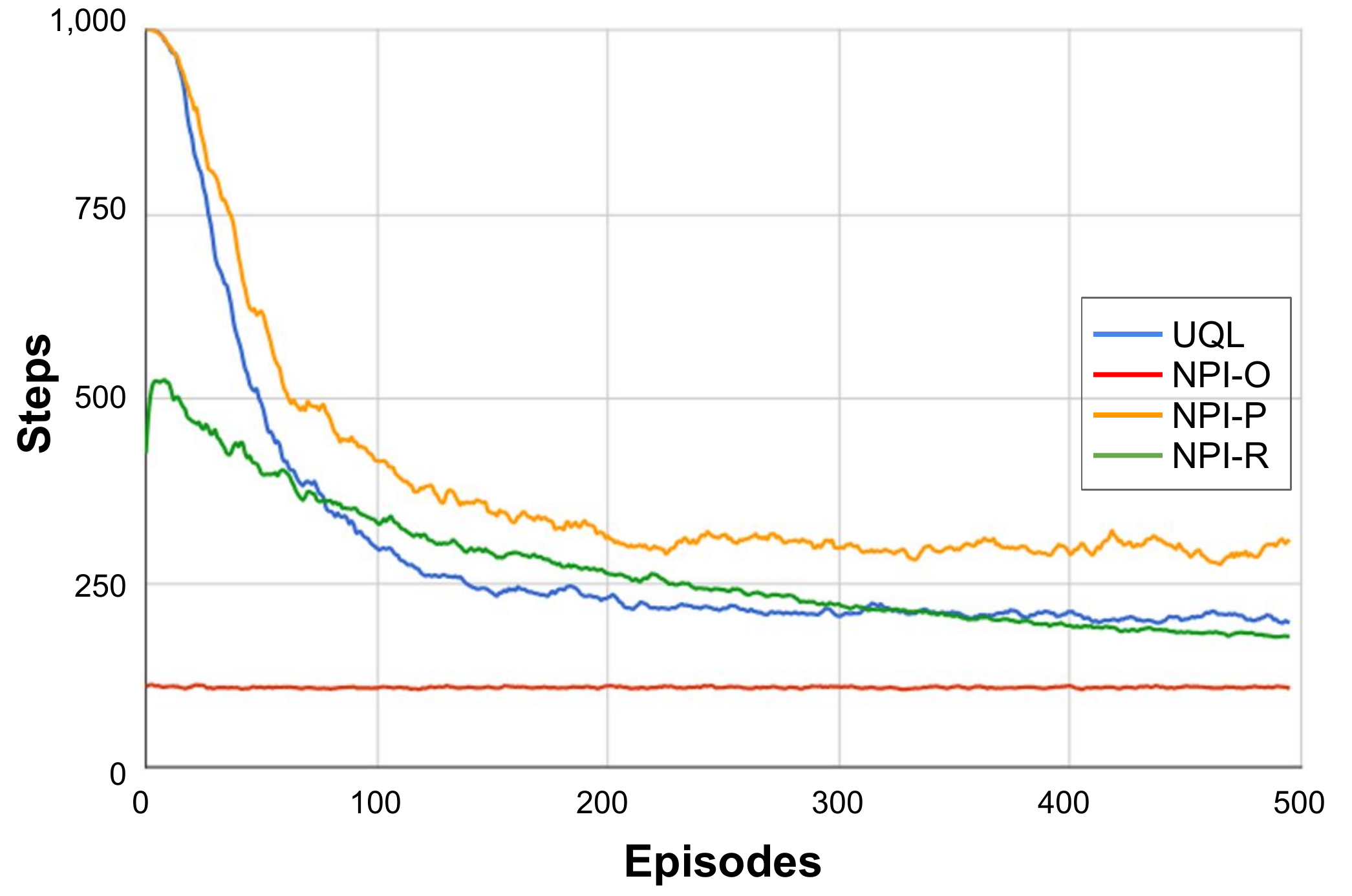}
\label{c7:fig:informative_persistent_c}}
\subfloat[Persistent informative]{\includegraphics[width=0.5\linewidth]{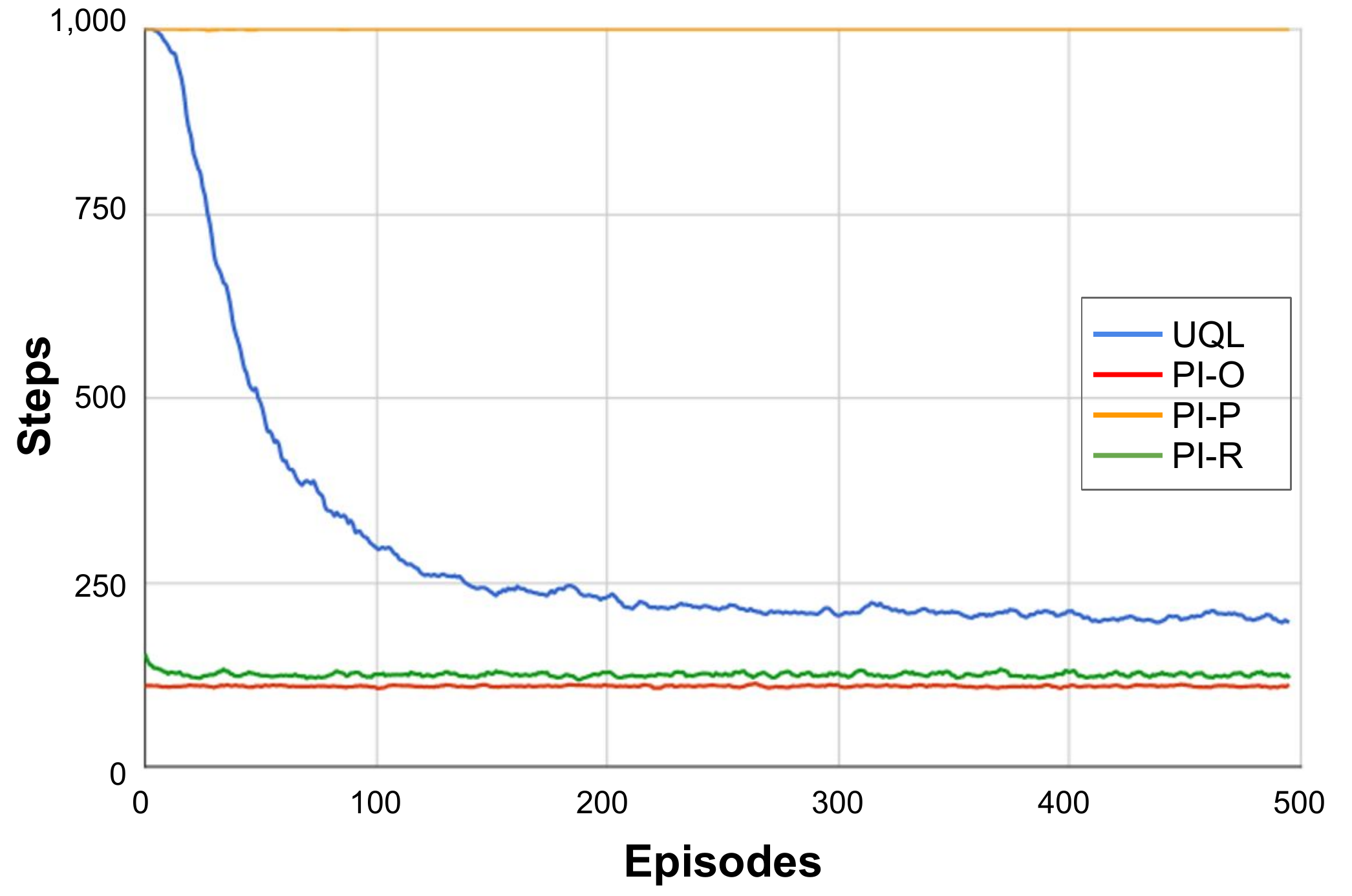}
\label{c7:fig:informative_persistent_d}}
\\
\caption{Three agents using informative advice are assisted by three different simulated users, initialized with either \textbf{O}ptimistic, \textbf{R}ealistic, or \textbf{P}essimistic values for accuracy and availability.
An unassisted Q-learning agent (UQL) is used as a benchmark. 
The figure shows that the persistent agents learn in fewer steps in comparison to the non-persistent agents when assisted by sufficiently accurate users.
}
\label{fig:state-based}
\end{figure}

For the mountain car environment, Figures~\ref{c7:fig:informative_persistent_c} and~\ref{c7:fig:informative_persistent_d} show the performance of IntRL agents, both non-persistent and persistent, using advice from three users with different levels of advice accuracy and availability. 
The advice that agents receive is an action recommendation, informing them of which action to take in the current state. 
When either agent, persistent or non-persistent, receives an action recommendation directly from the user on the current time step that action will be taken by the agent. 
The persistent agent will remember that action for the state it was received in, and use the PPR algorithm to continue to take that action in the future. 
Once the persistent agent has received an action recommendation from the user for a particular state, the user will not interact with the agent for that state in the future. 
The persistent agents require substantially fewer interactions than non-persistent agents. 
All persistent agents measured less than 1\% of steps with direct user interaction. 
Assuming a direct correlation between the number of interactions and the time required to perform those interactions, the use of persistence offers a large time reduction for assisting users. 
Results show that the number of interactions is much less for broad-advice agents compared to the state-based agents, allowing similar performance with much less effort from the trainer.

\begin{figure}
\centering
\subfloat[Steps per episode.]{\includegraphics[width=0.7\linewidth]{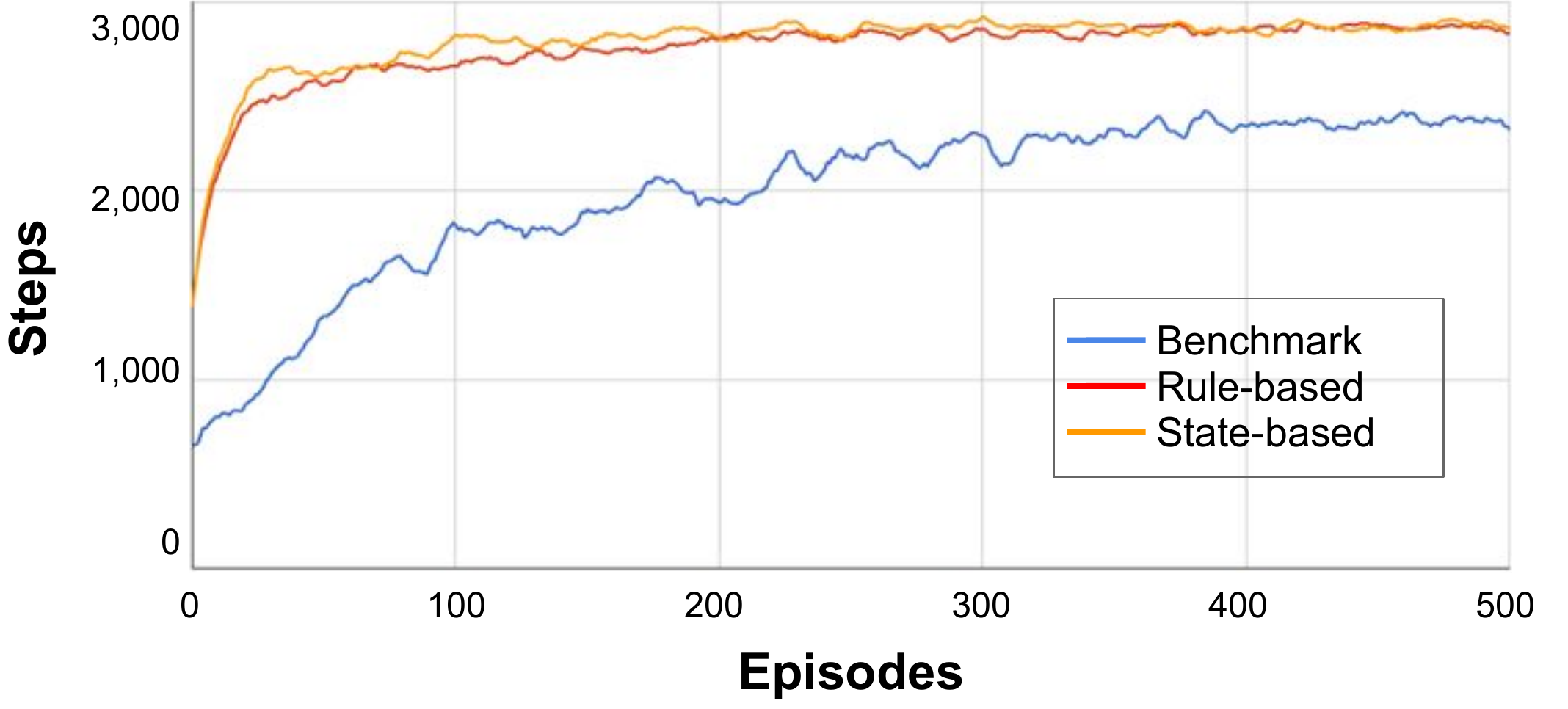}
\label{fig:sdc_rdr_steps_a}} 
\\
\subfloat[Reward per episode.]{\includegraphics[width=0.7\linewidth]{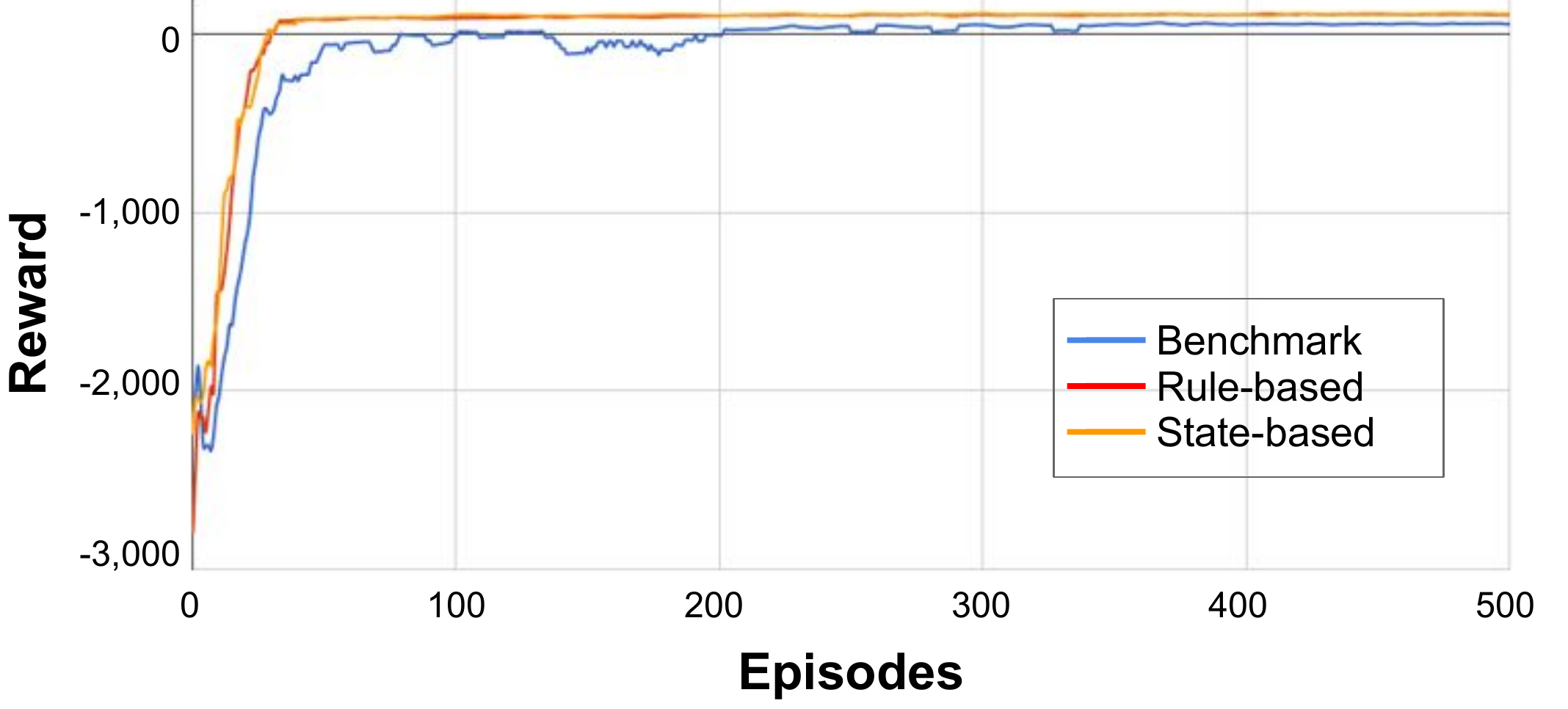}
\label{fig:sdc_rdr_reward_b}}
\caption{Steps and reward for state-based and rule-based (broad) IntRL agents for the self-driving car domain. 
Both agents obtained comparable performance, however, the advice required from the trainer is considerably less in the rule-based approach.} 
\label{fig:sdc_rdr}
\end{figure}

In the self-driving car environment, the aim of the agent is to avoid collisions and maximize speed. 
In the experiments, we created two simulated users to provide state-based and rule-based (broad) advice. 
Both agents outperformed the unassisted Q-Learning agent, both achieving a higher step count and reward. 
The obtained steps and reward are shown in Figure~\ref{fig:sdc_rdr_steps_a} and Figure~\ref{fig:sdc_rdr_reward_b} respectively.
Although the agent was forcibly terminated when it reached 3000 steps, Figure~\ref{fig:sdc_rdr_steps_a} shows that the agents never reached the 3000 step limit.
This is because the agents are given a random starting position and velocity at the beginning of each episode, some of which result in scenarios where the agent cannot avoid a crash.
Although both assisted agents outperformed the unassisted agent, between the state-based and rule-based (broad) methods, there is no considerable difference since both run a similar number of steps and collected a similar reward.
However, the number of interactions when using the state-based advice approach is 232 and when using broad advice is only 2.

\begin{figure}
    \centering
    \includegraphics[width=0.75\linewidth]{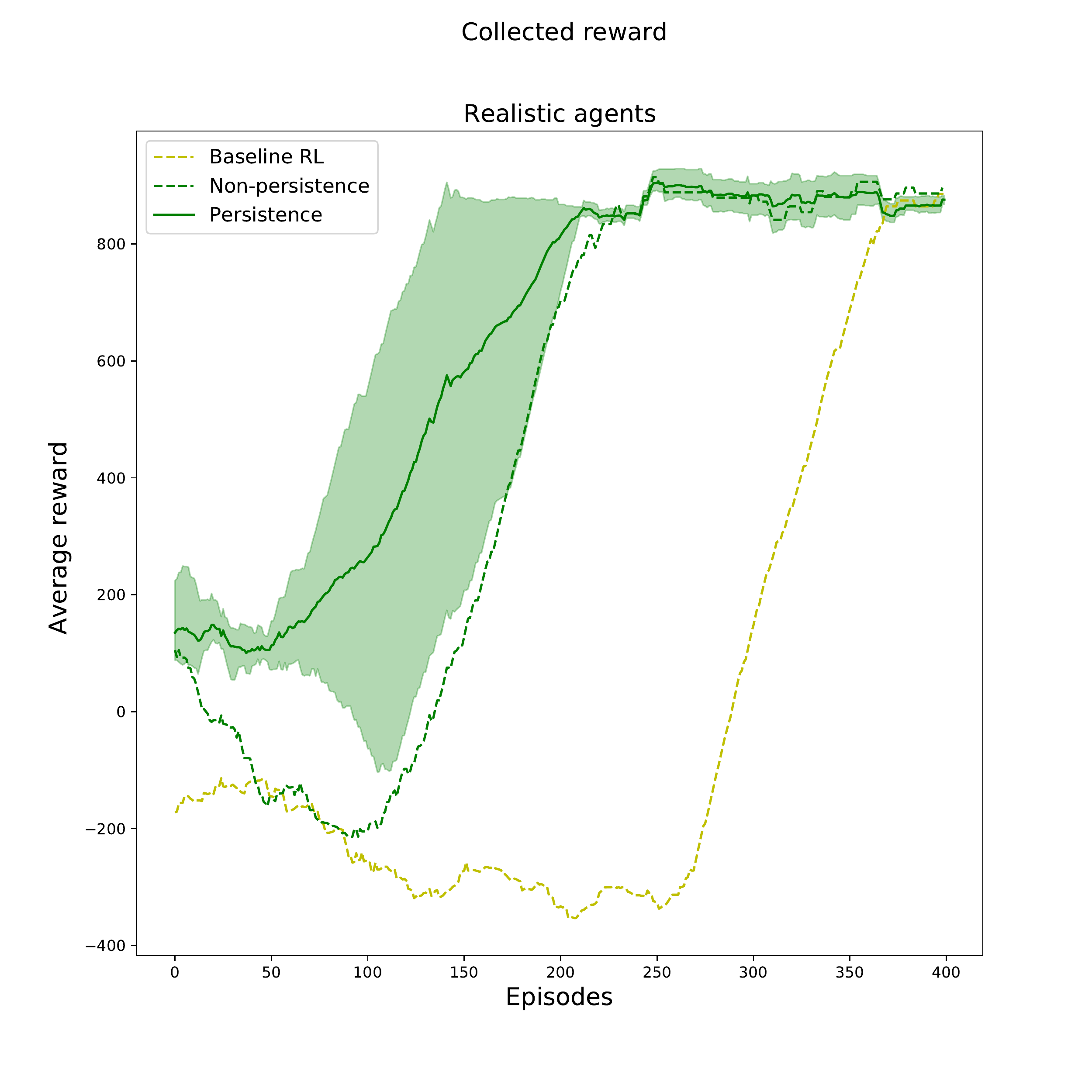}
    \caption{Result for deep reinforcement learning with an autonomous agent, non-persistent agent, and persistent agent built with Webots domestic robot environment.}
    \label{fig:53}
\end{figure}

Finally, for the simulated robot environment, the results obtained are shown in Figure~\ref{fig:53}. 
Non-persistent IntRL agent is shown by a green dashed line while the persistent IntRL agent is shown by the green solid line. 
Baseline RL is drawn with a yellow line used for benchmarking. 
Both agents supported by the trainer obtain better results than baseline RL. 
However, the persistent agent achieves convergence results slightly earlier than its non-persistent counterpart. 

\section{Conclusions}
In this work, we have introduced the concept of persistence in interactive reinforcement learning. 
Current methods do not allow the agent to retain the advice provided by assisting users. 
This may be due to the effect that incorrect advice has on an agent's performance. 
To mitigate the risk that inaccurate information has on agent learning, probabilistic policy reuse was employed to manage the trade-off between following the advice policy, the learned policy, and an exploration policy. 
Probabilistic policy reuse can reduce the impact that inaccurate advice has on agent learning.
Moreover, we present BPA, a broad-persistent advising approach to implement the use of PPR and generalized advice in continuous-state environments.

A more in-depth review of the generalization model is needed to get the best results for using PPR. 
The accuracy of the generalization model greatly affects the speed and convergence of the IntRL model.
Additionally, we suggest reducing the number of interactions with the trainer by reusing the action in the persistent model more often. 
When the agent reaches a new state that is already in memory, the agent reuses the recommended action immediately without interacting with the trainer. 
However, this should only be done when we have a good enough generalization model.


\bibliographystyle{ieeetr}
\balance
\bibliography{biblio}

\end{document}